\documentclass[letterpaper, 10 pt, conference]{ieeeconf}
\IEEEoverridecommandlockouts
\usepackage{moresize}
\usepackage{amsmath}
\usepackage{graphicx}
\usepackage{psfrag}
\usepackage{algorithm,algpseudocode}
\usepackage{amssymb}
\usepackage{subfigure}
\usepackage{verbatim}
\usepackage{latexsym}
\usepackage[dvipsnames]{xcolor}
\usepackage{cite}
\usepackage{stmaryrd}
\usepackage{pifont}
\usepackage{textcomp}
\usepackage{gensymb}
\usepackage{colortbl}
\usepackage{multirow}
\usepackage{hyperref}

\usepackage{epstopdf}
\usepackage{cleveref}
\usepackage{tabto}
\usepackage{tikz}

\graphicspath{./images}
\DeclareGraphicsExtensions{.pdf,.eps, .jpg, .png, .PNG}

\title{\LARGE \bf Navigation Framework for a Hybrid Steel Bridge Inspection Robot}
\vspace{-15pt}
\author{Hoang-Dung Bui, Hung M. La, \textit{IEEE Senior Member}
\thanks{This work is supported   by the U.S. National Science Foundation (NSF) under grants NSF-CAREER: 1846513 and NSF-PFI-TT: 1919127, and the U.S. Department of Transportation, Office of the Assistant Secretary for Research and Technology (USDOT/OST-R) under Grant No. 69A3551747126 through INSPIRE University Transportation Center.  
The views, opinions, findings and conclusions reflected in this publication are solely those of the authors and do not represent the official policy or position of the NSF and USDOT/OST-R.
}

\thanks{The authors are with the Advanced Robotics and Automation (ARA) Lab, Department of Computer Science and Engineering, University of Nevada, Reno, NV  89557, USA.  Corresponding author: Hung La, email: hla@unr.edu.}

\thanks{The source code of our implementation is available at the ARA lab's github: \url{https://github.com/aralab-unr/ara_navigation}}
}

\begin{document}

\maketitle

\begin{abstract} 
Autonomous navigation is essential for steel bridge inspection robot to monitor and maintain the working condition of steel bridge. Majority of existing robotic solutions requires human support to navigate the robot doing the inspection. In this paper, a navigation framework is proposed for the robot created by our Advanced Robotics and Automation (ARA) lab, called ARA robot,  \cite{nguyen2020practical, bui2020control} to run on mobile mode. In this mode, the robot is required to cross and inspect all the available steel bars. In our proposed framework, there are two main components: construct a graph to represent the bridge structure, and determine of the shortest path going through all the graph's edges at least one. The components consist of three algorithms: Structure Segmentation - segment the steel bridge structures into clusters, Graph Construction - build a graph to represent the structure, and Variant of Open Chinese Postman Problem - generate a shortest inspection path with any starting and ending points. Experiments on steel bridge structures setup highlight the effective performance of the proposed algorithms and the potential to apply to the ARA robot to run on real bridge structures.  We released our source code in Github for the research community to use.
\end{abstract}

\section{Introduction}\label{S.1}
Within the field of steel bridge structure monitoring, there is a significant attention of developing novel robots in the recent years. The influence of environment and transportation such as rain, solar radiation, overloading, friction degrades the bridge continuously. Thus, monitoring and maintenance are essential to ensure the bridge operation safety. The tasks can be done manually, however, it is time consuming, labor intensive, dangerous, affect to the traffic, and sometimes inaccessible for human in complex structures. For the reasons, there are varieties of robotic platforms \cite{seo2012tank, nguyen2020practical, wang2017analyses, Nguyen_IROS2019} developed to support human to do the task. These robots are magnetic-based that help them traverse on multiple angles of steel bridge structures. However, most of the robots are controlled manually by an operator.

\begin{figure}[ht]
    \centering
    \setcounter{subfigure}{0}
    \subfigure[]{\includegraphics[height=0.3\linewidth, width=0.3\linewidth]{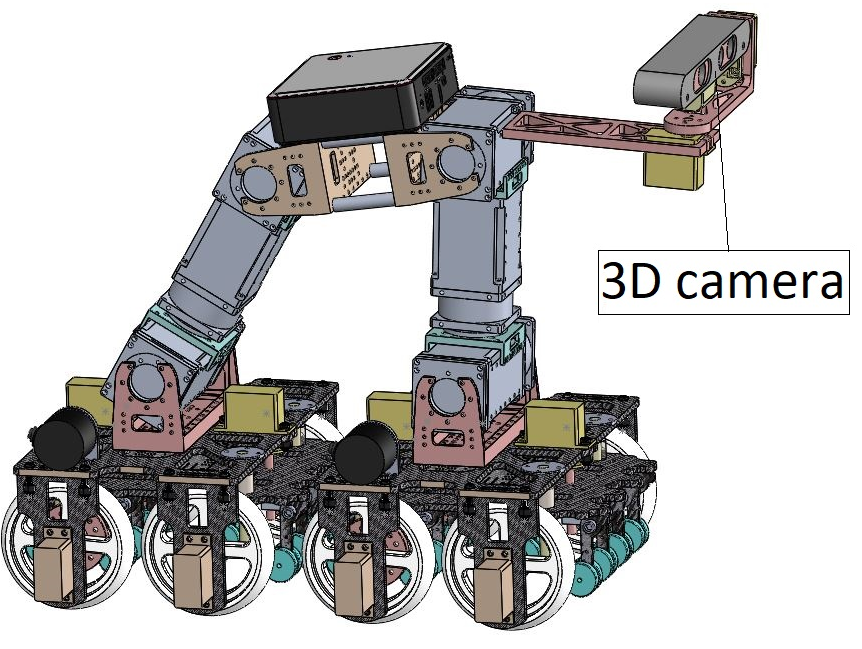}}
    \subfigure[]{\includegraphics[height=0.3\linewidth, width=0.3\linewidth]{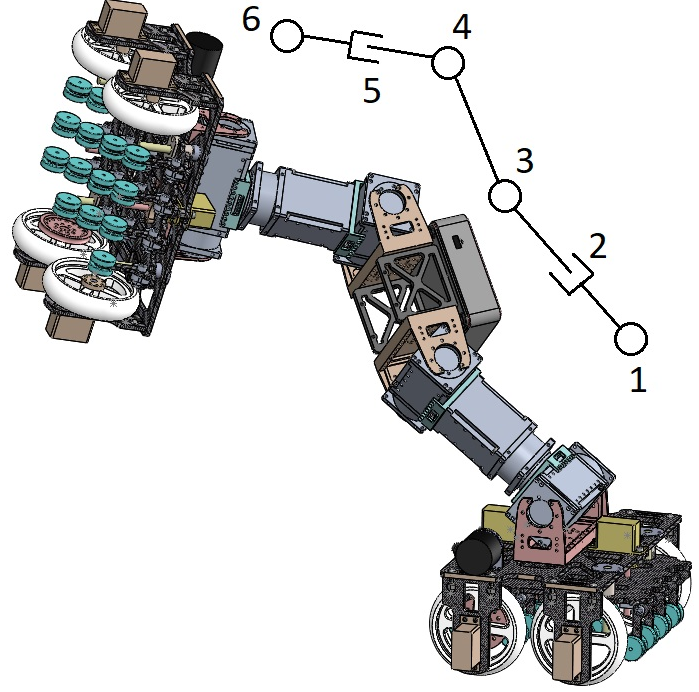}}
    \subfigure[]{\includegraphics[height=0.3\linewidth, width=0.3\linewidth]{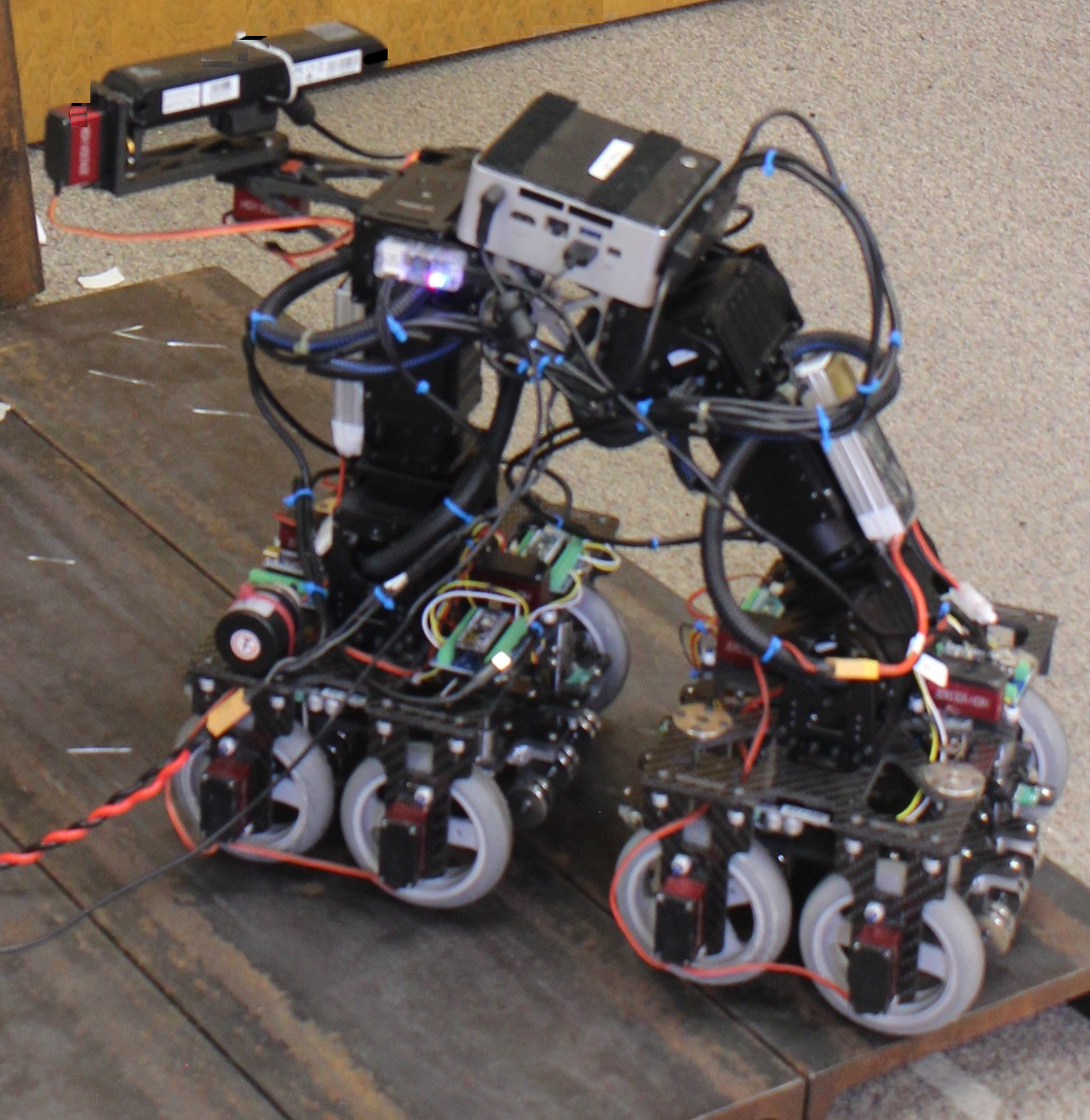}}
    \caption{ARA robot model in (a) \textit{mobile}, (b) \textit{inch-worm modes}, and (c) real robot}
    \label{fig:newModule}
\end{figure}

As an effort to go further in the field, the Advanced Robotics and Automation (ARA) lab has developed a bio-inspired hybrid robot - ARA robot \cite{nguyen2020practical, bui2020control} (Fig. \ref{fig:newModule}) with the aim to inspect the steel bridge structure autonomously. The robot is able to work in two modes: (1) mobile to traverse on smooth steel surface, and (2) inchworm - to change/jump to another steel bar surfaces. To continue the work of design and dynamic analysis \cite{nguyen2020practical} and control \cite{bui2020control},  in this paper a navigation framework is proposed for the robot to move easily and autonomously on smooth steel surface with the input data relying the on-board sensor.

\begin{figure}[ht]
    \centering
    \setcounter{subfigure}{0}
    \subfigure[]{\includegraphics[height=0.25\linewidth, width=0.46\linewidth]{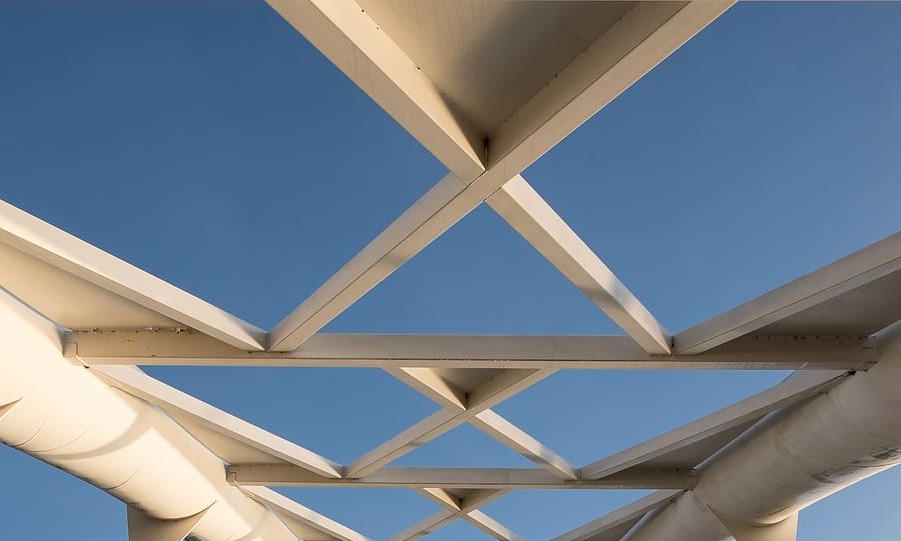}}
    \subfigure[]{\includegraphics[height=0.25\linewidth, width=0.46\linewidth]{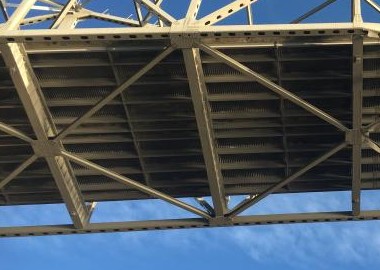}} \\
    \subfigure[]{\includegraphics[height=0.25\linewidth, width=0.46\linewidth]{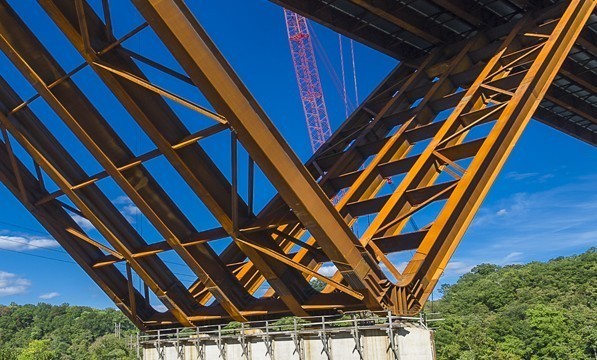}}
    \subfigure[]{\includegraphics[height=0.25\linewidth, width=0.46\linewidth]{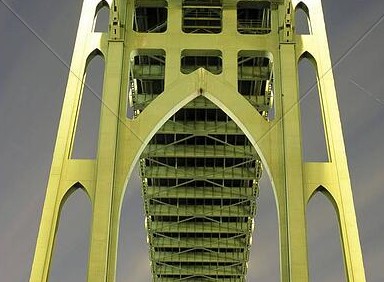}}
    \caption{The typical steel bridges structure: (a) cross shape, (b) K-Shape, (c) L-Shape, (d) and combination of shapes}
    \label{fig:TypeStruct}
\end{figure}

To move on smooth steel surfaces, the robot needs to navigate itself on varieties of structures in steel bridge as shown in Fig. \ref{fig:TypeStruct}, which consists of popular structures as \textit{Cross-}, \textit{T-}, \textit{I-}, \textit{K-} and \textit{L-} shape. The structure complexity and varied dimensions make motion planning task challenging. 
The motion planning needs a perception of the steel bridge structure to implement and a method to utilize limited workspace.
The bridge data can be provided offline, however, to improve the operation flexibility of the robot, the navigation system relies on the on-board sensor data, which is collected as robot operating. 
The on-board data is collected from the depth sensor, which was equipped on the robot. The sensor is only able to process a portion of the bridge in a view frame, thus, to navigate the robot to inspect the bridge continuously, the navigation system needs combine all the view frames and build a path going through all of them. To overcome the challenge, we propose a navigation framework that based on the idea: the bridge is divided into as a set of structures, each structure corresponding to a sensor frame is represented as a graph; then a variant algorithm of open Chinese Postman Problem (VOCPP) is developed to determine the shortest path to inspect all available steel bars on the structure with difference of starting and ending points.

The paper's structure is arranged as follows: Section \ref{Sec:RelWork} discusses the related works and the extension of this research compare to them. Section \ref{Sec:NaviFrame} describes the proposed navigation architecture for the ARA robot on mobile operation mode. The experimental results on several steel bridge structure prototypes are highlighted in section \ref{Sec:ExpRes}. Section \ref{Sec:Con} discusses the primary results and suggests the next steps for future research developments.

\section{Related Work}\label{Sec:RelWork}

There is a number of work related to the navigation of inspection robot for steel bridges  \cite{kotay1996navigating, xie2011edge, pagano2017approach}. In \cite{kotay1996navigating} particularly for autonomous steel bridge inspection robot, the authors proposed a task-level primitive and online navigation combining with IR sensor, which helps the robot move in local area. In \cite{xie2011edge}, the authors proposed a method to detect the features such as edges, walls, and corners on a large surface from laser light and use the features to navigate the robot. The method in \cite{pagano2017approach} supported the small size robot to move in a large-inside steel bridge space. In these researches, the robot's size were small comparing to the workspace - the surfaces which the robots ran on. The ARA robot, on the other hand,  was designed with manipulator form and carried several testing equipment's, thus its size is significant bigger than the mentioned robots. The circumstance requires a new navigation framework for the ARA robot, which can utilize the workspace better and localize the robot optimally to perform inspection tasks.

An important feature for steel bridge inspection robots is the limitation of the steel bar's dimensions, and small workspace for the robots to make a motion, especially for the situation of the ARA robot. The approximate methods in \cite{deits2015computing, jatsun2017footstep, hildebrandt2019versatile} built nice convex regions, which were represented by a set of curves. The methods made the construction of configuration space easy, however, reduced the workspace size and could make the robot motion infeasible. To overcome workspace's size reduction, in this paper a segmentation algorithm is proposed to separate the workspace into multiple clusters and represent them by a set of boundary points. 
The algorithm utilizes all the possible areas in the steel bridge structures, thus increases the probability of finding a path for the robot. With the irregular shape of the steel bars, Expectation Maximization - Gaussian Mixture Model (EM-GMM) method \cite{reynolds2009gaussian, pfeifer2019expectation, nguyen2012fast, blekas2005spatially} is utilized to segment the input data into a cluster set.

When perceiving the steel bridge structure, the navigation system represents it as a graph by its point cloud (PCL) data. Estimation of PCL features gets a significant attention \cite{goforth2020joint, gandler2020object, kraemer2017simultaneous}. These researches worked on objects that took small space comparing to the view of the depth sensors. The estimated features were the objects pose, shape or moving direction. In our research, the bridge is much larger than a sensor view frame. One steel bar can go from one side to another side of the sensor's frame. To estimate the features of the data, we developed a graph construction algorithm to represent the steel bridge structure as a graph in each sensor frame. In the best of the author's knowledge, we believed this is the first work dealing with this kind of PCL features estimation. 

From the built graph, the next step is to determine a shortest path to inspect all graph's edges, that is called \textit{inspection route}  or \textit{Chinese Postman Problem} (CPP) \cite{edmonds1973matching}. In CPP, a mailman is asked to travel all the streets in an area to deliver mails. He is required to find the shortest path to travel all the edge at least one. If he starts and ends in the same point, it is the original CPP. If the starting and ending points are different, that is open CPP. If there is a path, that goes through all the street exact on time, the graph is called Euler circuit and Euler path corresponding the CPP and the open CPP, respectively. There are different in our context. The real bridge is too long for the depth sensor to collect data in one frame, thus the bridge is separated into multiple frames. As the robot finishes inspection in one frame, it moves to the next one to continue the task. The ending point is selected to make the next inspection convenient. Therefore, the starting and ending points are different. Moreover, the varieties of the input steel bridge structures made the built graph become Euler circuit, Euler path or not any of them.
There are a number of researches working on CPP problem \cite{thimbleby2003directed, eiselt1995arc, ahr2006tabu}, which solved several aspects of CPP such as the street is directed, multiple mail-mans. However, there is no work can convert any graph with any starting and ending point into a Euler path. Therefore, we propose the VOCPP algorithm, which determines the shortest path in the current structure.

Particularly, our main contributions are:  
\begin{itemize}
    \item A navigation framework that helps the ARA robot be able to navigate on smooth surfaces on steel bridge structures;
    \item A graph construction algorithm that builds a graph to represent the steel bridge structure from the PCL data;
    \item A \textit{VOCPP} algorithm that finds a shortest path for the robot to inspect all steel bars in the graph with different starting and ending points;
    \item A segmentation algorithm that divides the steel bridge structure into steel bars and cross area regardless any kinds of input structure (tested on \textit{T-}, \textit{K-}, \textit{I-}, \textit{L-} and \textit{Cross-} shapes).
\end{itemize}

\section{Navigation Framework}\label{Sec:NaviFrame}
\subsection{Overall Architecture}
As the ARA robot works on \textit{mobile operation mode}, it traverses all steel bars of a steel bridge to detect the failure. To do the task in shortest time, it is needed to solve an optimization problem to determine the shortest path, which goes through all the available steel bars at least once. The task is sent to the motion planning module, and the inputs are PCL data, target position, and robot location from SLAM.
The PCL data of the steel bridge is filtered and then projected into a 2D plane, which is the \textit{xy} plane of the robot coordinate frame \cite{bui2020control}. The first main task of the motion planning is to build a graph to represent the steel bridge structure. The graph is built by two algorithms: 
structure segmentation - \textit{Algorithm 1} and graph construction - \textit{algorithm 2}. 
Algorithm 1 separates the processed data into the clusters, which are then classified as cross-area or steel bars. The cluster's boundaries are estimated by the Non-Convex Boundary Estimation (NCBE) algorithm \cite{bui2020control}. \textit{Algorithm 2} processes the boundaries data to directly make a graph. \textit{Algorithm 3} - VOCPP algorithm solves the optimization problem to find the shortest path. The VOCPP algorithm can solve any graph with any starting and ending points.
A path planning such as Rapidly-exploring Random Tree (RRT) receives the cluster boundaries and shortest path as inputs and builds a traverse path for the ARA robot to go along the steel bars. In this path planning, a simple algorithm called \textit{Point Inside Boundary Check - PIBC} is developed to determine efficiently the collision and availability of the robot configuration by the boundaries.

\begin{figure}[htb!]
    \centering
    \includegraphics[height=0.78\linewidth]{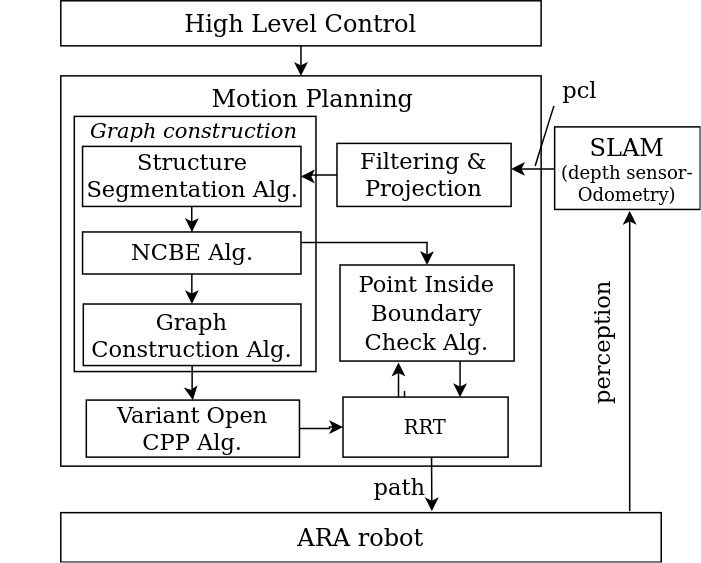}
    \caption{The proposed navigation framework on mobile mode}
    \label{fig:naviFrame}
\end{figure}

The output of the motion planner is sent to the ARA robot controller, which regulates the robot at the lower level to perform the motion. The framework is shown in Fig. \ref{fig:naviFrame}.

\subsection{Graph Construction}
Representing a steel bridge structure as a graph is a critical component of the navigation system. To build the graph, the input PCL data is processed through three algorithms: Structure Segmentation, NCBE, and Graph Construction as shown in Fig. \ref{fig:naviFrame}. NCBE algorithm is presented detail in  \cite{bui2020control}, thus the others will be discussed in the following.

\noindent \textbf{Structure Segmentation Algorithm}:
Classifying the cross-area and steel bars is critical to determine the types of the steel bridge structures. Therefore, steel bridge perception is a requirement to build a graph. The bridge structure can be perceived by a convolutional neutral network (CNN) \cite{narazaki2018automated}, however, it requires a huge cost to make a dataset, especially for steel bridges, which are usually located in rugged terrain. Moreover, with the variety of the structures, in the best of the author's knowledge, there is no CNN working on steel bridge structure detection.

To detect the steel bridge structure, we propose a method based on the bridge's geometric features that they typically consist of two components: (1) steel bars and (2) cross areas. As the method segments the structure into these two components, the bridge structure is represented by a graph whose edges and vertices are the steel bars and cross areas, respectively. 
The steel bars are irregular in shape with one dimension being much longer than other, and their two ends connect two cross areas. The cross areas are regular in shape and their dimensions are similar, and there are at least two steel bars connected with. The difference in the geometric features of the steel bars and cross areas is used to classify them.

From the feature analysis, an algorithm is developed based on EM-GMM classification  \cite{reynolds2009gaussian, nguyen2012fast, blekas2005spatially}. The EM-GMM classification is able to separate the structure into multiple clusters with irregular shapes. However, it requires a specific cluster number as input, and if there are several types of distribution in the data, EM-GMM does not works well. To deal with unknown cluster numbers, the algorithm works on a set of cluster numbers and determines which is the most suitable number $n_o$ for cluster segmentation. The number $n_o$ is then inputted back into the EM-GMM algorithm that generates a set of clusters. 

\begin{figure}[ht]
    \centering
    \setcounter{subfigure}{0}
    \subfigure[]{\includegraphics[height=0.35\linewidth, width=0.45\linewidth]{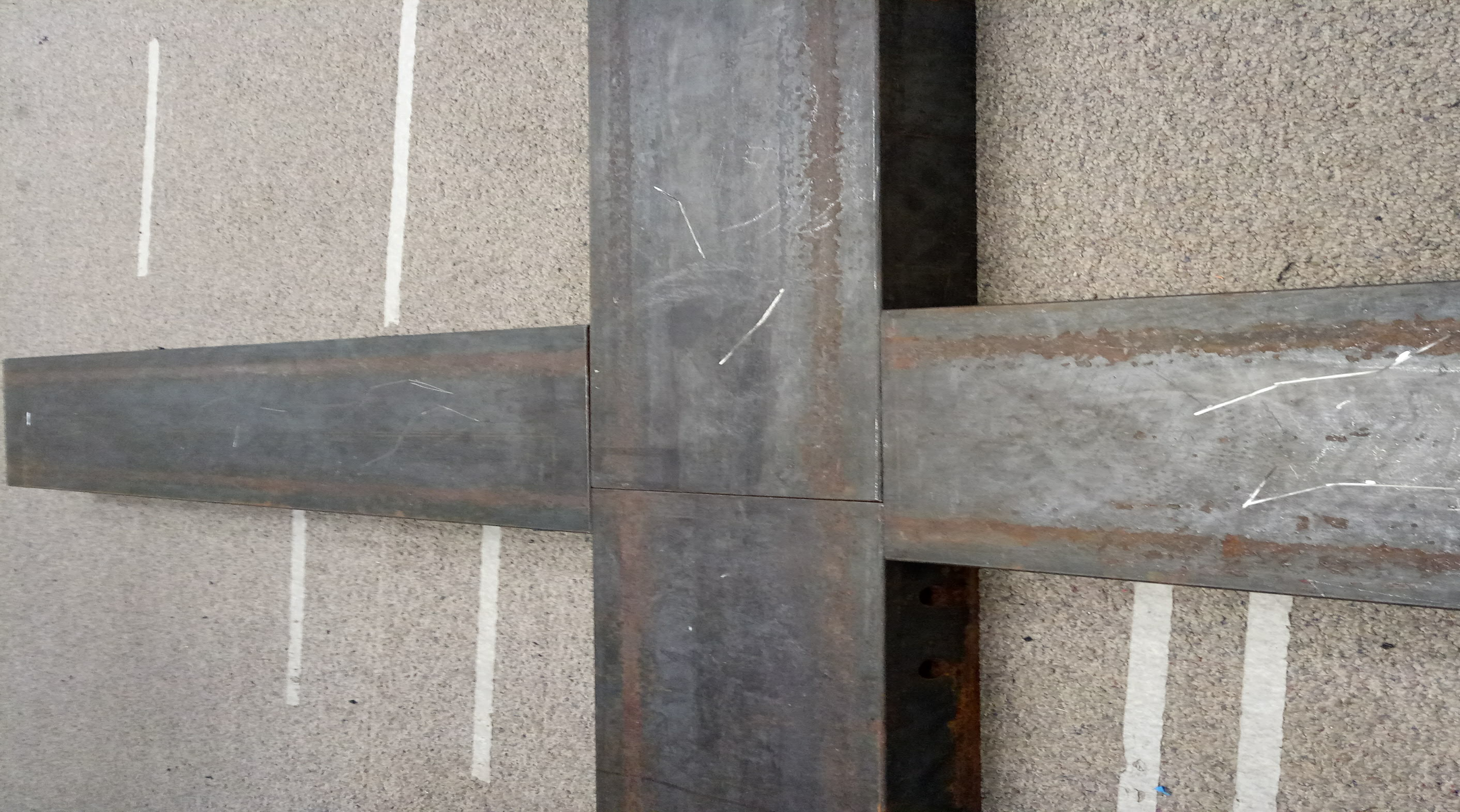}}
    \subfigure[]{\includegraphics[height=0.38\linewidth, width=0.52\linewidth]{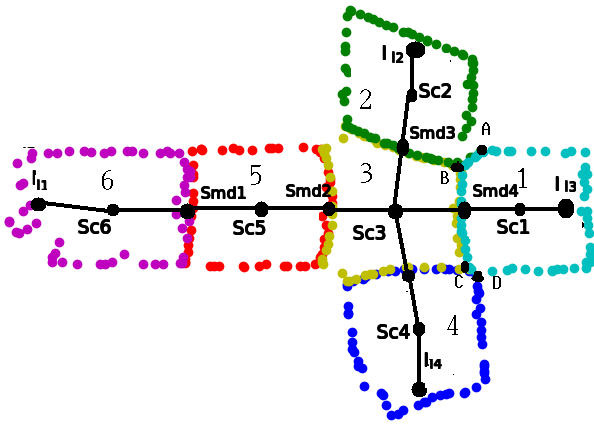}}
    \caption{(a) A \textit{cross-} shape steel bar and (b) its segmentation}
    \label{fig:bar_cross}
\end{figure}

To find the best cluster number for the steel bridge segmentation, a new concept - neighbor cluster - is introduced.  \textit{Two clusters are considered as neighbors if they share a border with the length at least $l_b$}. From that, in Fig. \ref{fig:bar_cross}b, cluster 1 and 3 are considered neighbors because the border \textit{BC} is longer than the threshold $l_b$. Although there are some contacting points, there can be no neighbor-relationship between clusters 1 and 2 because the border \textit{AB} is shorter than $l_b$. The same applies to cluster 1 and 4 since \textit{CD} $\leq l_b$. From this idea, if the cluster numbers $n_o$ are optimal, the cross area should be the one with most neighbor clusters $n_m$ as shown in Fig. \ref{fig:bar_cross}b (cluster 3). Therefore, the first idea is that the cluster number $n_o$ is optimal if it makes the cross area cluster having $n_m$ with highest value.

\setlength{\textfloatsep}{4pt}
\begin{algorithm}[ht]
\small
\caption{Structure Segmentation Algorithm}\label{alg:PclSeg}
    \begin{algorithmic}[1]
        \Procedure{Pcl Segmentation}{$P_{cl}, N_{cmin}, N_{cmax}$}
            \State Initialize $n_{index}, r_n=\emptyset$
            \For{$n_c \in \{ N_{cmin}, N_{cmax}\} $}
                \State $S_c$ = \textit{EM-GMM Algorithm}($P_{cl}, n_c$)
                \For {$j \in \{0,n_c\}$} \label{step:ncbest}
                    \State $b[j] = NCBE(S_c[j], sliding\_factor)$
                \EndFor \label{step:ncbeend}
                \State {Determine neighbor clusters for $S_c[j]$} \label{step:neighClus}
                \State{Get the most and second most neighbor numbers $n_m, n_s$}
                \State{Calculate $r$ from Eq. \ref{Eq:optclus}, and add to set $r_n$}
                \State{Add the $n_c$ in set $n_{index}$}
            \EndFor
            \State{Select the highest $r$ in $r_n$, and get $n_o$ from $n_{index}$} \label{step:r}
            \State {$S_{b}$ = EM-GMM Algorithm($P_{cl}, n_o$)} \\
            \Return $S_{b}$
        \EndProcedure
    \end{algorithmic}
\end{algorithm}

The most neighbor cluster numbers $n_m$ works well for \textit{K-}, \textit{T-}, \textit{Cross-} shape. However, for the structures such as \textit{I-} and \textit{L-} shapes, the highest number of neighbor is not enough to segment correctly. For \textit{L-}shape, as the cluster number is more than three, there are several clusters with the same neighbor cluster number, and it is not possible to find the cross area cluster. For \textit{I-}shape, there is two cross-area clusters in the structure. Thus, it is needed another feature to handle these shapes, and the difference between the most neighbor number $n_m$ and the second most neighbor number $n_s$ with the same $n_c$ is considered. Combining the two features, the ratio $r$ is defined in Eq. (\ref{Eq:optclus}), and the highest $r$ will be selected. 

\begin{align} \label{Eq:optclus}
    r = \frac{n_m}{n_m + n_s} + \frac{n_m}{n_c}.
\end{align}
In Eq. (\ref{Eq:optclus}), the first term tends to keep $n_c$ small, and the second term tends to make $n_m$ large. $n_c$ is the number of clusters.

The procedure detail is presented by the psudo-code in \textit{Algorithm \ref{alg:PclSeg}}. The inputs are the PCL data, and a range of cluster number from $N_{cmin}$ to $N_{cmax}$. As $n_c$ runs in the range of [$N_{cmin}$ to $N_{cmax}$], EM-GMM algorithm segments the PCL data into a set of clusters $S_c$. From line \ref{step:ncbest} to \ref{step:ncbeend}, NCBE algorithm determines the set of boundaries $b$ corresponding to $S_c$. The neighbor number for each cluster is determined in line \ref{step:neighClus}, and from that the most and second most cluster numbers are specified. From $n_c, n_m$, and $n_s$, the ratio $r$ is calculated, then the $r$ and $n_c$ are putted into the sets $r_n$ and $n_{index}$, respectively. Step \ref{step:r} selects the highest $r$, which is inputted into EM-GMM algorithm to get the most appropriate segmented clusters $S_{b}$.

The clusters set $S_b$ are sent to NCBE algorithm, which returns the boundary set of the clusters $S_{bo}$. $S_{bo}$ is then inputted to the graph construction algorithm with a length threshold $d_{min}$ for cluster neighborhood. 
\begin{algorithm}[ht]
\small
\caption{Graph Construction Algorithm} \label{alg:GraphEsti}
    \begin{algorithmic}[1]
        \Procedure{Graph Construction}{data cluster set $S_{bo}$, threshold $d_{min}$}
            \State Initialize $E = \{\}, V = \{\}$.
            \State{Calculate the center point set $S_c$ of boundary set $S_{bo}$} \label{step:cenPoint2}
            \State{Determine neighbor matrix $M_n$} \label{step:Neighmatrix}
            \State{Determine borders set $S_{bd}$ among the neighbors from $S_{bo}, M_n$}
            \State{Determine the middle point set $S_{md}$ of the border set $S_{bd}$} \label{step:middlepoint}
            \State{Add all vertices $S_c, S_{md}$ into $V$: $V = S_c \cup S_{md}$} \label{step:vertices}
            \State{Build edges $e$ by connecting vertices in $V$ if the corresponding clusters are neighbors (from $M_n$}) \label{step:edges}
            \State{Estimate line set $S_l$ to fit the cluster set $S_c$ by $PCA$} \label{step:fitlines}
            \State{Determine intersection of feature lines \& their boundaries: $I_l = S_l \cap S_b$} \label{step:edgePoint}
            \For{i $\in I_l$} \label{step:checkandAdd}
                \For{j $\in V$}
                    \If{ (distance from $I_l[i]$ to $V[j]$) $>$ $d_{min}$}
                        \State{Add $I_l[i]$ into V}
                        \State{Edge $e$ = from $I_l[i]$ to its center vertex}
                        \State{Add $e$ into $E$}
                    \EndIf
                \EndFor
            \EndFor \label{step:endfor1}\\
            \Return $G = (V,E)$
        \EndProcedure
    \end{algorithmic}
\end{algorithm}

\noindent \textbf{Graph Construction Algorithm}: From the clusters and their boundaries, this algorithm (\textit{Algorithm \ref{alg:GraphEsti}}) builds the graph by determining the edges $E$ and nodes $V$ from the cluster boundaries set $S_{bo}$ (from NCBE Algorithm) and a distance threshold $d_{min}$. Firstly, the center points $S_c$ of clusters, neighbor matrix $M_n$ - determine the neighbor relationship among clusters, borders set $S_{bd}$, and border middle points $S_{md}$ are determined from step \ref{step:cenPoint2} to \ref{step:middlepoint} (Fig. \ref{fig:bar_cross}).
b
\begin{algorithm}
\small
\caption{Variant Open CPP}\label{alg:PathVisiAllEdge}
    \begin{algorithmic}[1]
        \Procedure{Minimal Path}{$G = (\mathcal{V}, \mathcal{E}), v_s, v_t$}
            \State Initialize odd vertices set $S_{ov}=\{\}$
            \State Find all odd vertices and add to $S_{ov}$.
            \If{$S_{ov} = \emptyset$} \label{step:eulerian}
                \State{Find shortest edges set $\mathcal{E}_e $ connecting $v_s$ and $v_t$
                \State Assign $\mathcal{G}_n = (\mathcal{V}, \mathcal{E} \cup \mathcal{E}_e)$}, go to \ref{step:solEuPa}.
            \EndIf
            \If{($v_s, v_t$) $\subset$ $S_{ov}$} \label{step:bothinside}
                \State{$S_{mov} = S_{ov}\backslash \{v_s, v_t \}$, $G_m = G$, go to \ref{step:contoEulPath}.}
            \EndIf
            \If {$v_s \notin$ $S_{ov}$ and $v_t \in$ $S_{ov}$} \label{step:vtinside}
                \State{select a vertex $v_c \in S_{ov}$, which creates the shortest path $\mathcal{E}_c$ connect $v_s$ to it}
                \State {$S_{mov} = S_{ov}\backslash \{v_t, v_c \}$, $\mathcal{G}_m = (\mathcal{V}, \mathcal{E} \cup \mathcal{E}_c)$, go to \ref{step:contoEulPath}}
            \EndIf
            \If{$v_s \in$ $S_{ov}$ and $v_t \notin$ $S_{ov}$} \label{step:vsinside}
                \State{select a vertex $v_c \in S_{ov}$, which creates the shortest path $\mathcal{E}_c = $ connect $v_t$ to $v_c$}
                \State{$S_{mov} = S_{ov}\backslash \{v_s, v_c \}$, $\mathcal{G}_m = (\mathcal{V}, \mathcal{E} \cup \mathcal{E}_c)$, go to \ref{step:contoEulPath}}
            \EndIf
            \If{ $v_s \notin S_{ov}$ and $v_t \notin S_{ov}$} \label{step:allout}
                \State{select two vertices $v_{c1}, v_{c2} \in S_{ov}$, which creates two paths $\mathcal{E}_{d1}, \mathcal{E}_{d2}$ that sum of them is shortest}
                \State{$S_{mov} = S_{ov}\backslash \{v_{c1}, v_{c2} \}$, $G_m = \{\mathcal{V}, \mathcal{E} \cup \mathcal{E}_{d1} \cup \mathcal{E}_{d2}\}$, go to \ref{step:contoEulPath}.}
            \EndIf
            \State \label{step:contoEulPath} Using $\mathcal{G}_m$, find a set of edges $\mathcal{E}_a$ whose sum are shortest to convert all vertices in $S_{mov}$ to even vertices.
            \State{$\mathcal{G}_n = (\mathcal{V}, \mathcal{E}_m \cup \mathcal{E}_a)$ }
            \State \label{step:solEuPa} Find the Eulerian path $P_{robot}$ ($v_s \longrightarrow v_t$) from $\mathcal{G}_n$ \\
            \Return $P_{robot}$
        \EndProcedure
    \end{algorithmic}
\end{algorithm}
All the points in $S_c$ and $S_{md}$ are added into the vertices set $V$. The edge set is built by $M_n$ and $V$ in step \ref{step:edges}. After that, a set of feature's lines that are fit to the boundaries are calculated by \textit{PCA} method \cite{fukunaga2013introduction}. Step \ref{step:edgePoint} determines the intersection set $I_l$ of the feature's lines and their cluster boundary $S_{bo}$.
From step \ref{step:checkandAdd} to \ref{step:endfor1}, there are dual loops to check whether the distance from a point in $I_l$ to a vertex in $V$ is longer than the threshold $d_{min}$. If yes, then that point and an edge, which connects it to its center point, are added into the vertex set $V$ and the edge set $E$, respectively. The algorithm outputs $G = (V,E)$, which is sent to \textit{Algorithm} \ref{alg:PathVisiAllEdge}.


\subsection{Variant Open CPP - VOCPP Algorithm}
The built graph is sent to VOCPP algorithm to determine the shortest path to inspect all the available steel bars. There are multiple steel bridge structures, thus the input graph could consist of Euler circuit, Euler path or none. Algorithm \ref{alg:PathVisiAllEdge} - VOCPP is proposed to solve the optimization problem, which is based on \textit{Euler Theorem} \cite{edmonds1973matching} for Eulerian trail. The algorithm converts any input graph into an \textit{open CPP} graph by adding the shortest path into the input graph and make a new one. The added path is selected by using Dijkstra's algorithm \cite{cormen2009introduction}. The output is the shortest path, which visits each edge at least one, and starts and ends at the predefined positions.
The \textit{pseudo-code} is shown in Algorithm \ref{alg:PathVisiAllEdge}. The inputs are the graph $G=\{V,E\}$, starting and target vertex $v_s,v_t$, respectively. Step \ref{step:eulerian} checks whether an Eulerian circuit exists in the graph. If yes, a path generated by Dijkstra's algorithm with two vertices $v_s, v_t$ is added into the graph. After that, the algorithm will jump to step \ref{step:solEuPa}. If the odd vertex set $S_{ov}$ are not empty, step \ref{step:bothinside} checks whether $v_s, v_t$ belong to the set. If yes, both are popped out, and a set of shortest paths is added to convert all the remained odd nodes in $S_{ov}$ to even. Steps \ref{step:vtinside}, \ref{step:vsinside}, and \ref{step:allout} check whether the vertex $v_s$ or $v_t$ stays in the $S_{ov}$, or if both are out of the set. If any input vertex is odd, it will be popped out. A vertex in $S_{ov}$ that possesses the shortest paths to the even input vertices is connected by a shortest path to convert the even input vertex into odd. After that, the selected vertex is also popped out of $S_{ov}$. All then go to step \ref{step:contoEulPath} to convert all the odd vertices in the remaining $S_{ov}$ into even vertices by Dijkstra's algorithm. Step \ref{step:solEuPa} will find the shortest path that traverses all edges of the graph by Fleury's algorithm \cite{thorup2000near}.

\begin{figure*}[h!]
    \centering
    \setcounter{subfigure}{0}
    \subfigure[]{\includegraphics[height =0.14\linewidth, width =0.17\linewidth, angle=90]{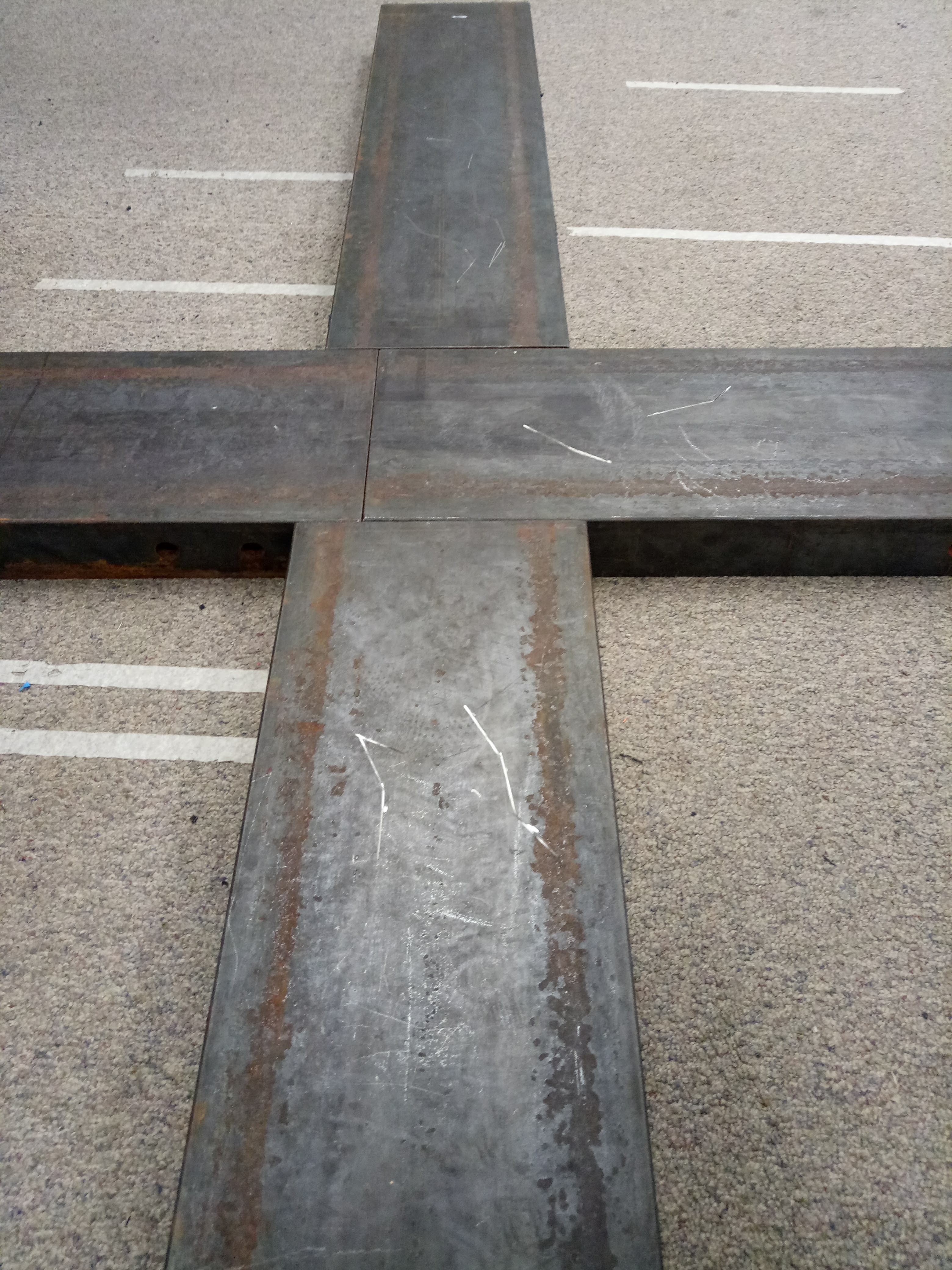}} \quad
    \subfigure[]{\includegraphics[height =0.14\linewidth, width =0.17\linewidth, angle=90] {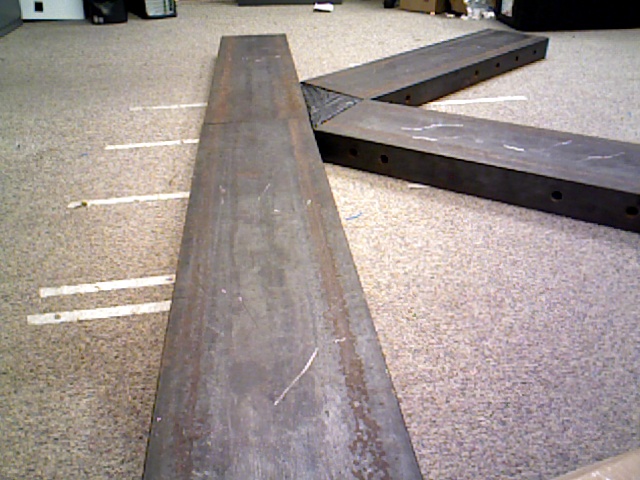}} \quad
    \subfigure[]{\includegraphics[height =0.14\linewidth, width =0.17\linewidth, angle=90] {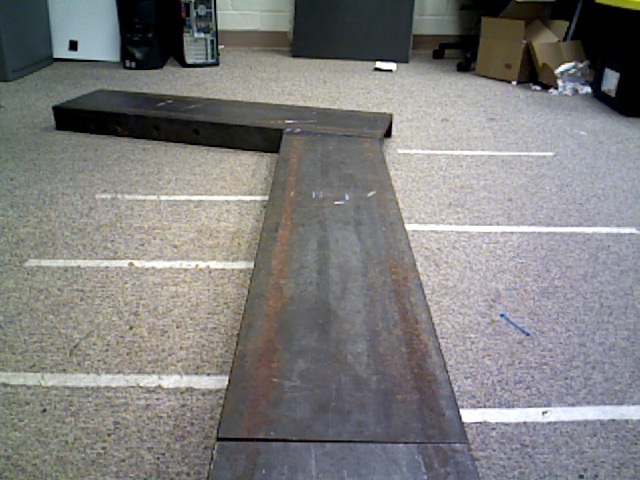}} \quad
    \subfigure[]{\includegraphics[height =0.14\linewidth, width =0.17\linewidth, angle=90] {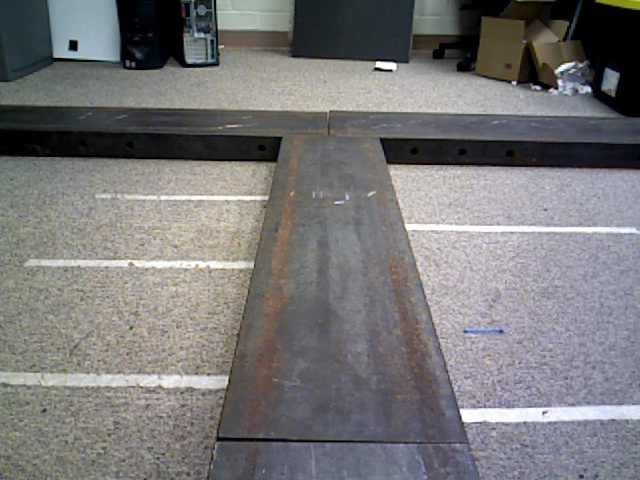}} \quad
    \subfigure[]{\includegraphics[height =0.14\linewidth, width =0.17\linewidth, angle=90] {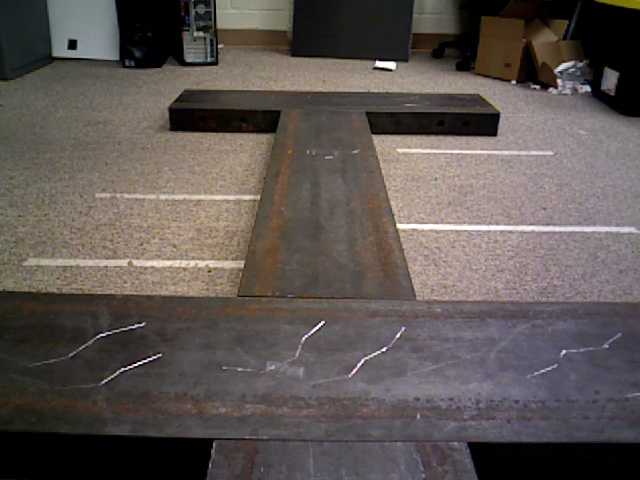}} \\
    \subfigure[]{\includegraphics[height =0.15\linewidth, width =0.17\linewidth]{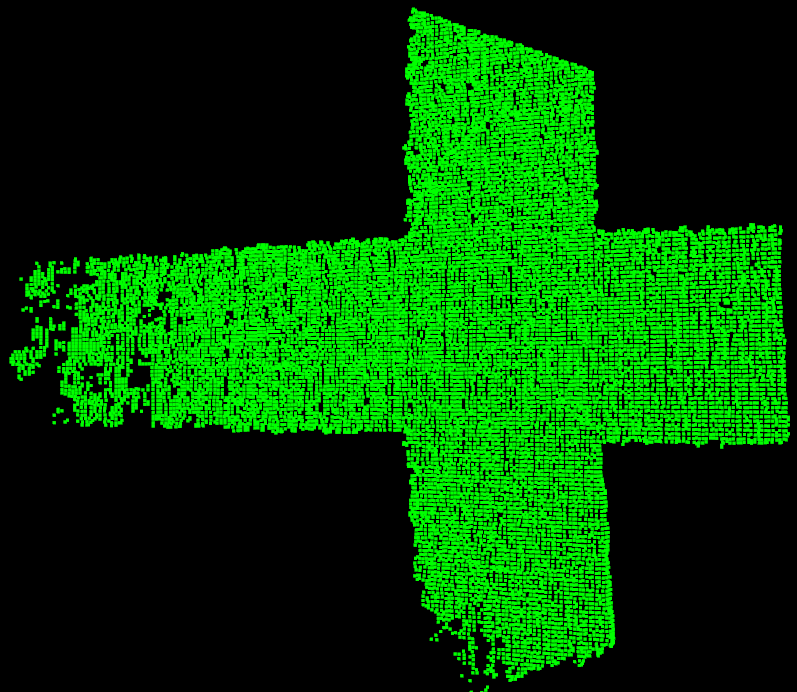}}
    \subfigure[]{\includegraphics[height =0.15\linewidth, width =0.17\linewidth]{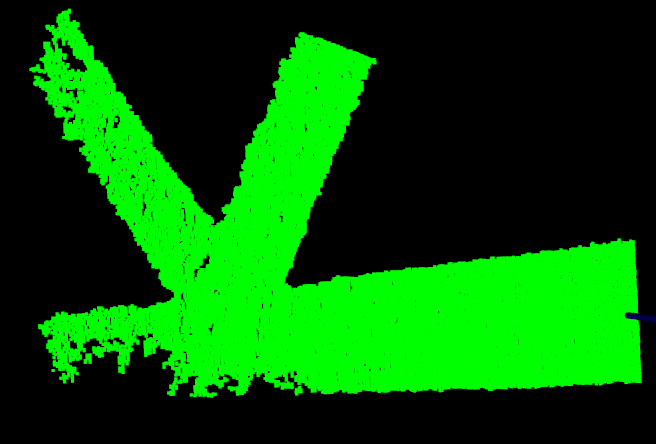}}
    \subfigure[]{\includegraphics[height =0.15\linewidth, width =0.17\linewidth]{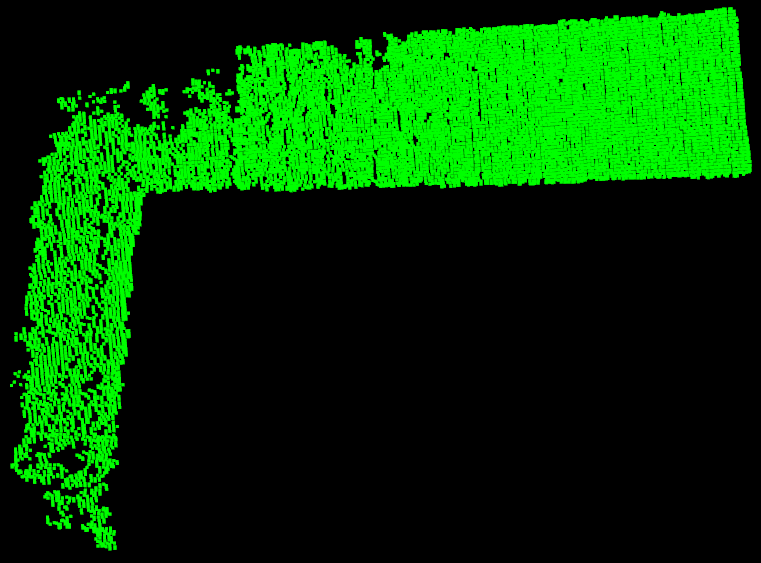}}
    \subfigure[]{\includegraphics[height =0.15\linewidth, width =0.17\linewidth]{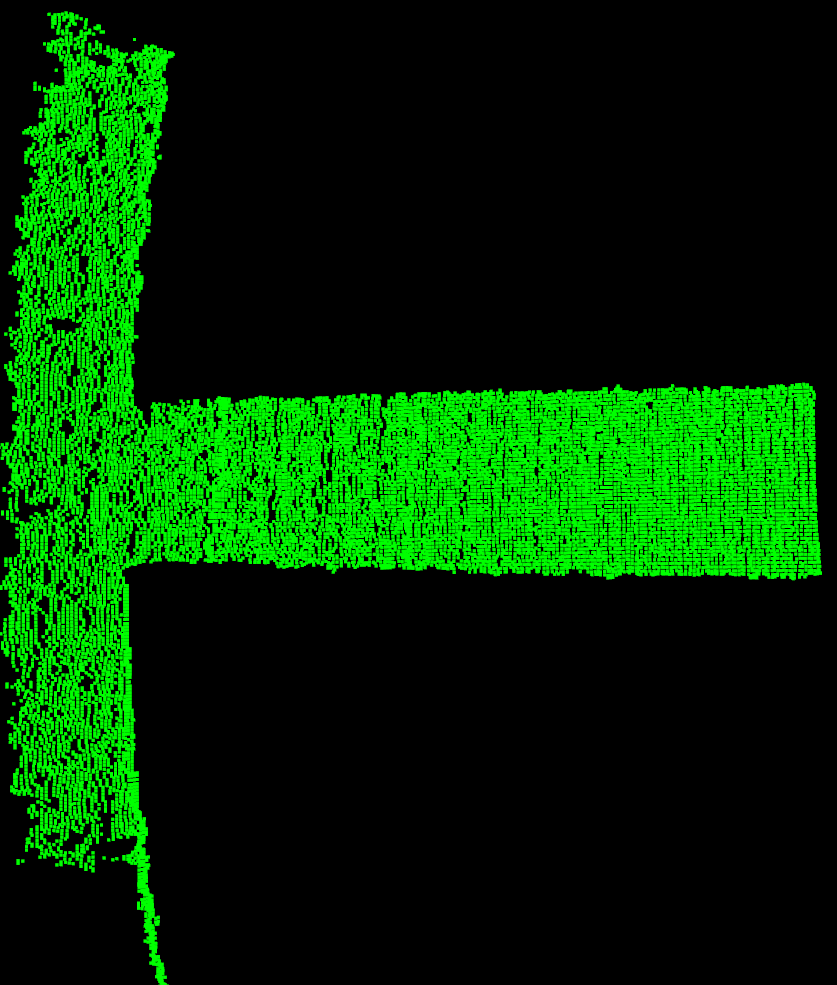}}
    \subfigure[]{\includegraphics[height =0.15\linewidth, width =0.17\linewidth]{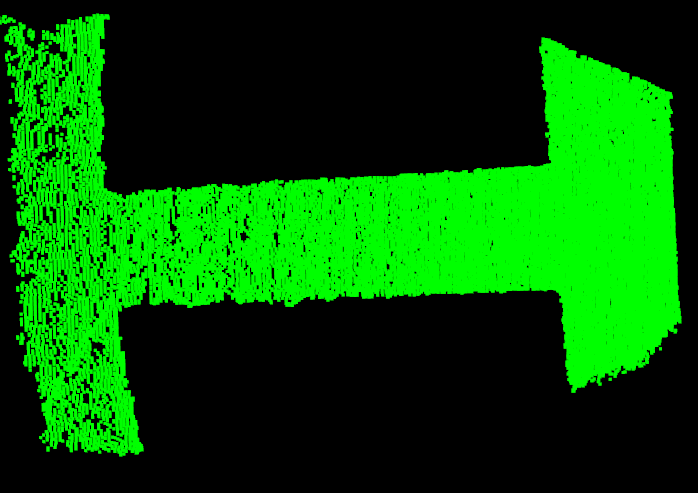}}\\
    \subfigure[]{\includegraphics[height =0.15\linewidth, width =0.17\linewidth]{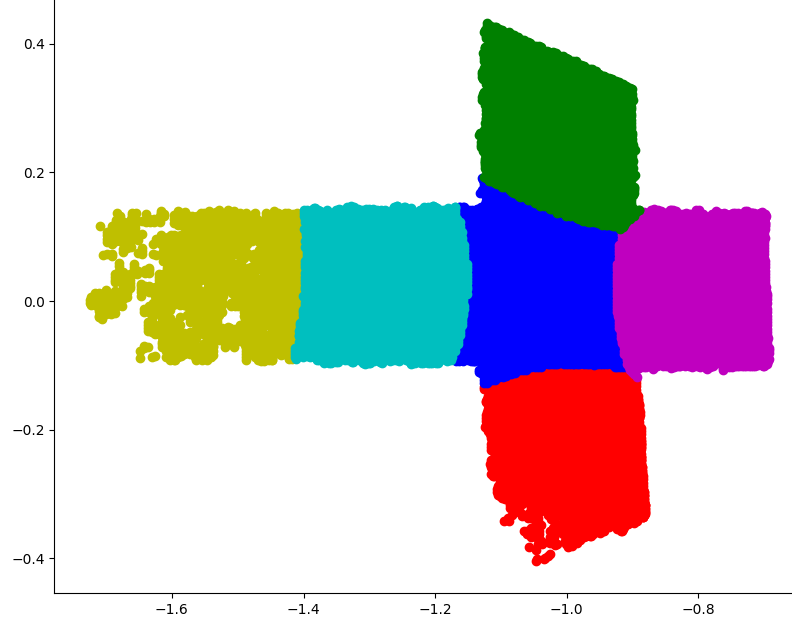}}
    \subfigure[]{\includegraphics[height =0.15\linewidth, width =0.17\linewidth]{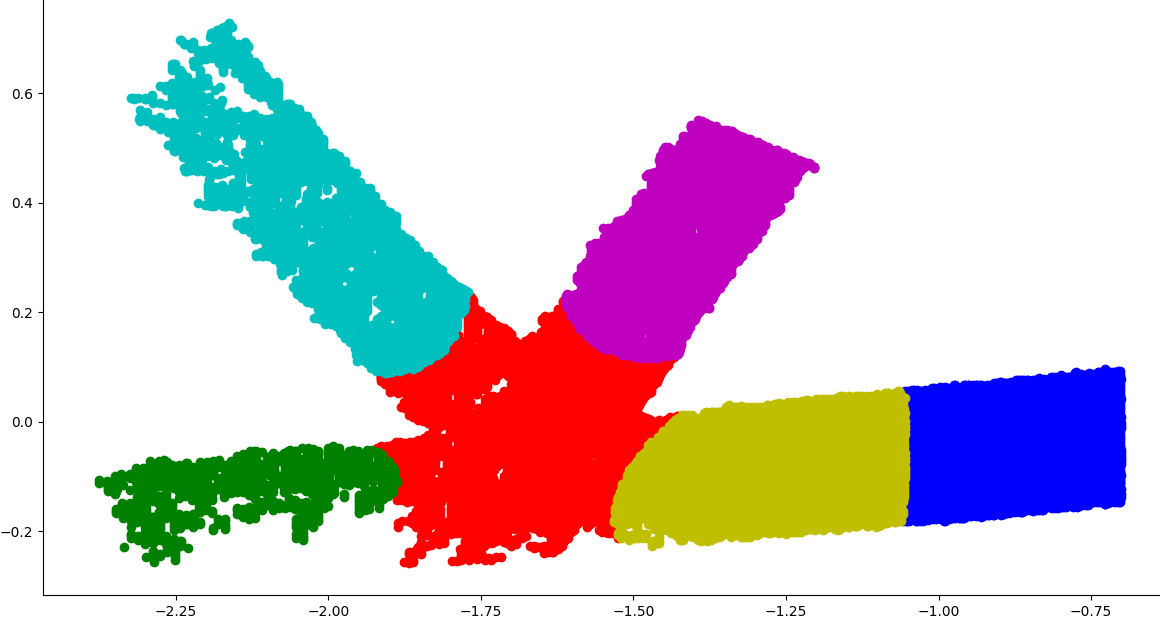}}
    \subfigure[]{\includegraphics[height =0.15\linewidth, width =0.17\linewidth]{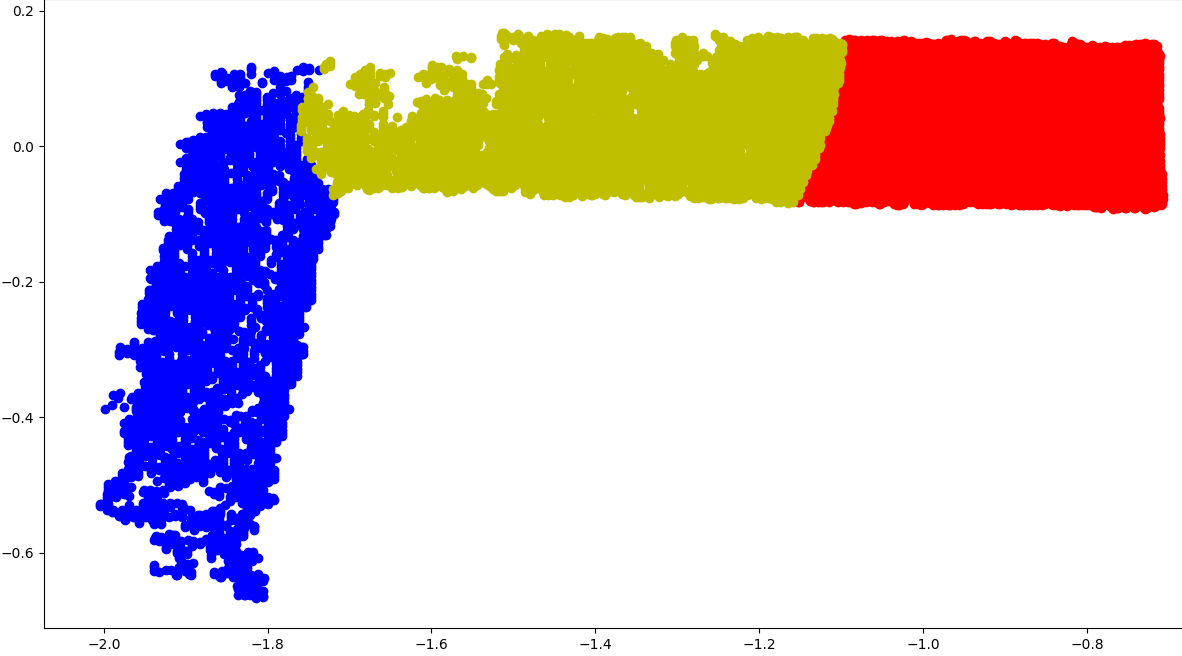}}
    \subfigure[]{\includegraphics[height =0.15\linewidth, width =0.17\linewidth]{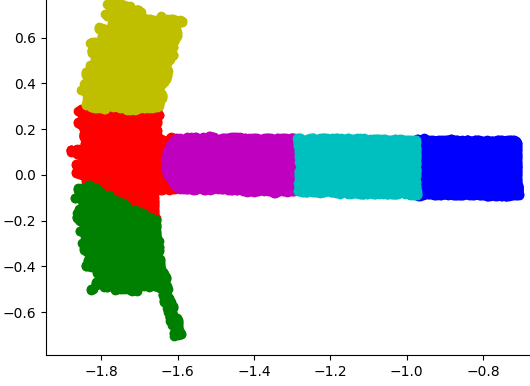}}
    \subfigure[]{\includegraphics[height =0.15\linewidth, width =0.17\linewidth]{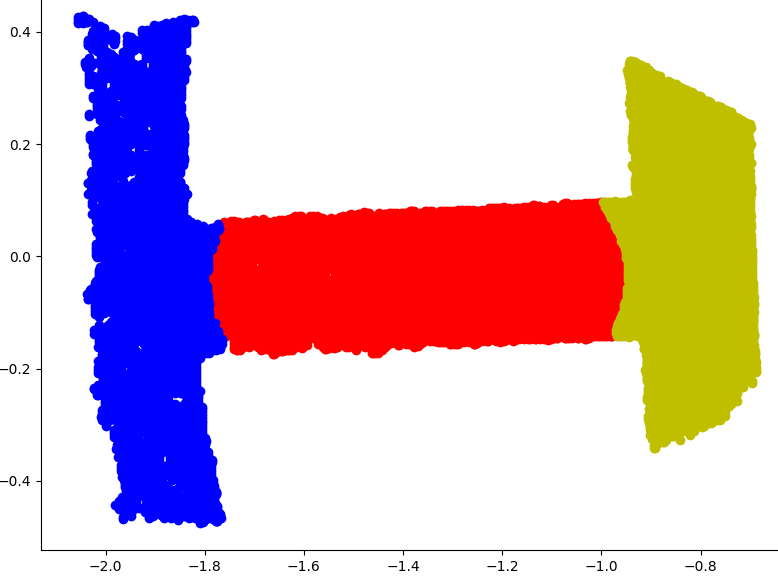}}\\
    \subfigure[]{\includegraphics[height =0.15\linewidth, width =0.17\linewidth]{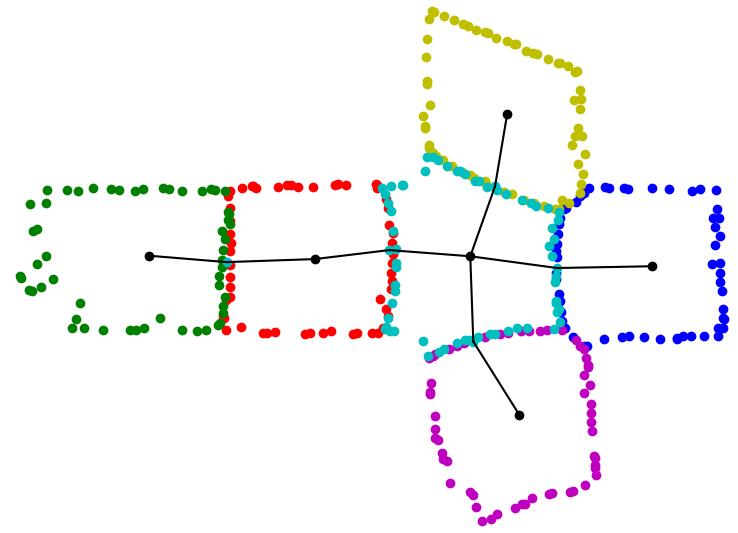}}
    \subfigure[]{\includegraphics[height =0.15\linewidth, width =0.17\linewidth]{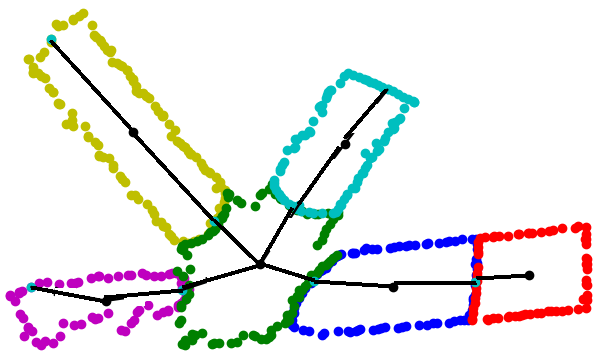}}
    \subfigure[]{\includegraphics[height =0.15\linewidth, width =0.17\linewidth]{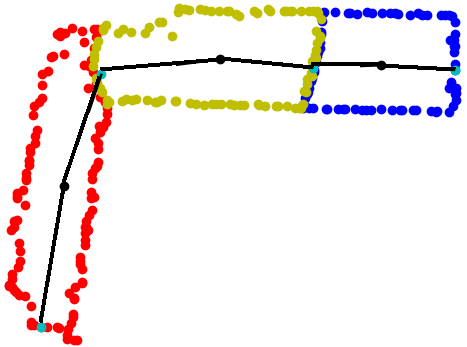}}
    \subfigure[]{\includegraphics[height =0.15\linewidth, width =0.17\linewidth]{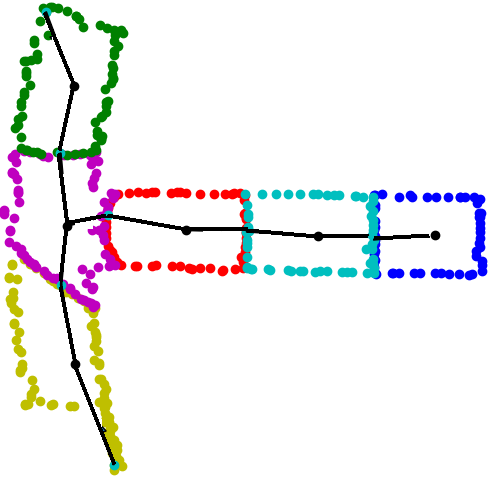}}
    \subfigure[]{\includegraphics[height =0.15\linewidth, width =0.17\linewidth]{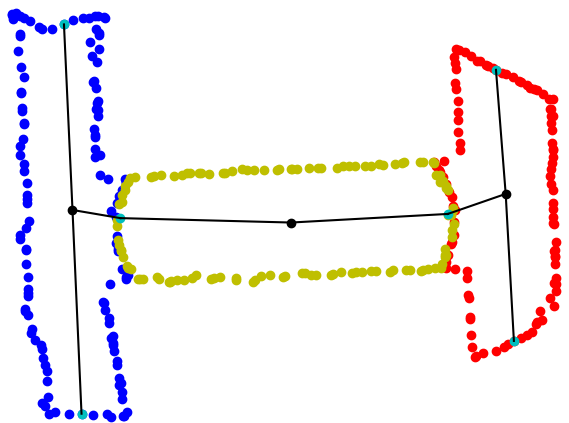}} \\
    \subfigure[]{\includegraphics[height =0.15\linewidth, width =0.17\linewidth]{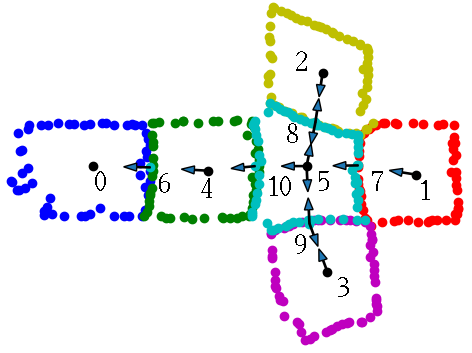}}
    \subfigure[]{\includegraphics[height =0.15\linewidth, width =0.17\linewidth]{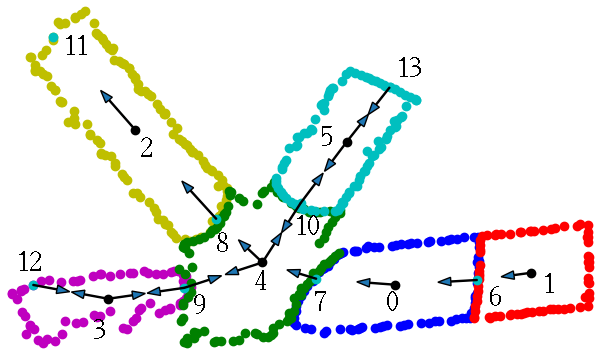}}
    \subfigure[]{\includegraphics[height =0.15\linewidth, width =0.17\linewidth]{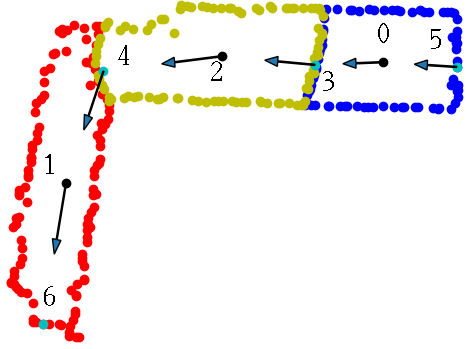}}
    \subfigure[]{\includegraphics[height =0.15\linewidth, width =0.17\linewidth]{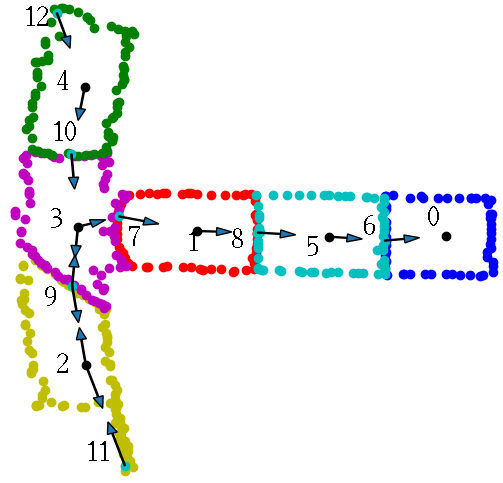}}
    \subfigure[]{\includegraphics[height =0.15\linewidth, width =0.17\linewidth]{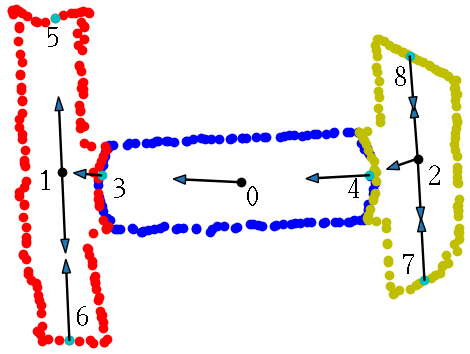}}
    \caption{The images of input structures (a-e), the corresponding point clouds (f-j), the segmentation (k-o), boundary estimation and graph construction (p-t), and the shortest path (u-y)}
    \label{fig:segment_boundary}
\end{figure*}

\section{Experiment Results}\label{Sec:ExpRes}
We implemented the experiments on the ARA robot depth sensor, which collected the PCL data on the setup of steel bridge structures such as \textit{L}-, \textit{Cross}-, \textit{T}-, \textit{K}, and \textit{I-} shape. Due on the lab condition, several steel bars were combined on the ground to make the shape types mentioned as shown in the first row of Fig. \ref{fig:segment_boundary}. The second row was the PCL data of \textit{Cross}-, \textit{K}-, \textit{L}-, \textit{T}-, and \textit{I}- shapes after filtering and projected into the robot coordinate frames. To present the experiment result succinctly, the bridge structure's images and their PCL data were rotated 90 degree with the viewpoint from the right to left. All the algorithms integrated into ARA-robot control program and ran on ROS. 

\subsection{Graph Construction}
The graph construction is shown in the third and fourth rows in Fig. \ref{fig:segment_boundary}.

\textbf{Structure Segmentation Algorithm}: The result of the segmentation algorithm was shown in the third row of Fig. \ref{fig:segment_boundary}. 
The algorithm was able to segment properly the cross areas and the steel bar parts, except the \textit{I}- shape. The reason was that there were two cross areas in \textit{I}-shape structure, which made \textit{algorithm \ref{alg:PclSeg}} confusing. This problem will be improved in the future research. By a robust graph construction algorithm, however, the graph of \textit{I}- shape could be still built and helped generate the path for the robot. In the cases of \textit{T-}, \textit{Cross-}, and \textit{L-} shapes, one of the steel bars is much longer than others, thus it was segmented into two portions.
After segmenting, the clusters were sent to the \textit{NCBE algorithm} \cite{bui2020control}, which gave back the boundaries of the clusters $S_{bo}$.

\textbf{Graph Construction Algorithm}: processed the boundary data from the \textit{NCBE algorithm}, and outputted the result on the fourth row. 
In the row, the graph was built based on the center point, edge points of the boundary and the border points between the neighbor clusters. The graphs covered the steel bars' length for most structures, except the \textit{Cross}- structure (Fig. \ref{fig:segment_boundary}p). It was because of the distance from the center points to the corresponding edge points are too close, the depth sensor could cover all the distance. Therefore, it was no need an edge to connect that two points. In Fig. \ref{fig:segment_boundary}v, the edge (12-3) does not go along with the steel bar but crossing on one edge. It occurred because the data quality is not good, and influencing the line fitting algorithm. The same reason happens to the \textit{T}- structure in Fig. \ref{fig:segment_boundary}s.

\subsection{VOCPP algorithm}
The figures in the last row of Fig. \ref{fig:segment_boundary} showed the shortest path generated by the \textit{VOCPP} algorithm. Due to the probabilistic feature of EM-GMM algorithm, the cluster indices changed each time running. Starting at a random vertex $v_s$, the robot follows the arrow lines to the next vertex, and comes back if there is a dead end. The route ends at the predefined ending vertex $v_t$, and the generated paths are optimal with shortest length. Again, due to the point cloud data quality, in Fig. \ref{fig:segment_boundary}v,x, the edges \textit{(3,12)}, \textit{(2,11)} are not possible for robot to traverse. To prevent the robot go on the edges, the motion planner needs to handle this case.


\section{Conclusion and Future Work}\label{Sec:Con}
A navigation framework was proposed for the ARA robot to run on mobile mode. In this mode, the robot was required to cross and inspect all the steel bars. The most significant contributions of this research were the navigation framework and three
algorithms, which could process the depth data, then outputted a traverse path for the robot. Algorithm 1 - Structure Segmentation segmented the steel bar structures into two sets: steel bars and cross areas. Based on the segmentation result, the graph construction - algorithm 2 built a graph that represented the bridge structure. From the graph, algorithm 3 - VOCPP generated a shortest path for the robot to move and inspect all the available steel bars. 

There are several aspects, which needs to be improved in our research. The efficiency of algorithms needs to extend to process more bridge structures. The stability of the segmentation algorithm, which depends on the probability method, is not in well-operating and outputs sometime inappropriate segmentation. The drawbacks will be solved in the next research, which can help the robot operate in fully autonomous navigation.
\bibliographystyle{unsrt}
\bibliography{RefFile}

\end{document}